%% file: main.tex
\documentclass{article}
\usepackage[preprint]{log_2024}			

\usepackage{booktabs}						
\usepackage{colortbl}
\usepackage{multirow}						
\usepackage{amsfonts}						
\usepackage{graphicx}						
\usepackage{duckuments}						
\usepackage{adjustbox}
\usepackage{svg}
\usepackage{longtable}
\usepackage{graphicx}

\usepackage[numbers,compress,sort]{natbib}	

\newtheorem{definition}{Definition}

\title[Topological Deep Learning with State-Space Models: A Mamba Approach for Simplicial Complexes]{Topological Deep Learning with State-Space Models: \\A Mamba Approach for Simplicial Complexes}

\author[M. Montagna et al.]{%
Marco Montagna\\
Sapienza University of Rome\\
\email{marco.montagna@uniroma1.it}\And
Simone Scardapane\\
Sapienza University of Rome\\
\email{simone.scardapane@uniroma1.it}\And
Lev Telyatnikov\\
Sapienza University of Rome\\
\email{lev.telyatnikov@uniroma1.it}
}

\begin{document}

\maketitle

\input{sections/0_abstract}

\input{sections/1_introduction}
\input{sections/2_related_works}
\input{sections/3_method}
\input{sections/4_experiments}
\input{sections/5_results}
\input{sections/6_conclusions}

\bibliographystyle{unsrtnat}
\bibliography{TopoMamba}

\newpage
\appendix
\input{sections/appendix}

\end{document}

%% file: sections/0_abstract.tex
\begin{abstract}
Graph Neural Networks based on the message-passing (MP) mechanism are a dominant approach for handling graph-structured data. However, they are inherently limited to modeling only pairwise interactions, making it difficult to explicitly capture the complexity of systems with $n$-body relations. To address this, topological deep learning has emerged as a promising field for studying and modeling higher-order interactions using various topological domains, such as simplicial and cellular complexes. While these new domains provide powerful representations, they introduce new challenges, such as effectively modeling the interactions among higher-order structures through higher-order MP.
Meanwhile, structured state-space sequence models have proven to be effective for sequence modeling and have recently been adapted for graph data by encoding the neighborhood of a node as a sequence, thereby avoiding the MP mechanism. In this work, we propose a novel architecture designed to operate with simplicial complexes, utilizing the Mamba state-space model as its backbone. Our approach generates sequences for the nodes based on the neighboring cells, enabling direct communication between all higher-order structures, regardless of their rank. We extensively validate our model, demonstrating that it achieves competitive performance compared to state-of-the-art models developed for simplicial complexes.

\end{abstract}

%% file: sections/1_introduction.tex
\section{Introduction}

Graph neural networks (GNNs) \cite{scarselliGraphNeuralNetwork2009} are widely used across various fields due to their ability to leverage the underlying structure of data. By aggregating features from neighboring nodes and edges, GNNs can learn effective representations that often lead to better performance compared to models that do not use structural information. This principle forms the foundation of Message Passing Models \cite{gilmerNeuralMessagePassing2017}, which is the approach employed by many of the most popular GNNs.
However, the limitations of these methods have been explored in previous works \cite{morrisWeisfeilerLemanGo2019}. A key drawback is that graphs inherently represent only pairwise interactions, which makes it difficult to capture more complex relationships between nodes. Additionally, GNNs face challenges in modeling interactions between distant nodes, as multiple message-passing steps are required for information to propagate over long distances \cite{liDeeperInsightsGraph2018}. \\
To try and overcome some of the GNNs limitations several efforts have been made to adapt Transformers \cite{vaswaniAttentionAllYou2023} for use with graph-structured data \cite{yunGraphTransformerNetworks2020, kimPureTransformersAre2022}. These approaches treat the graph as a complete graph, computing attention between all pairs of nodes to facilitate communication, regardless of the distance between them. However, this method has the significant drawback of the attention matrix scaling quadratically with the number of nodes, making it impractical for large graphs.
\begin{figure}[h]
    \centering
    \includegraphics[width=.9\textwidth]{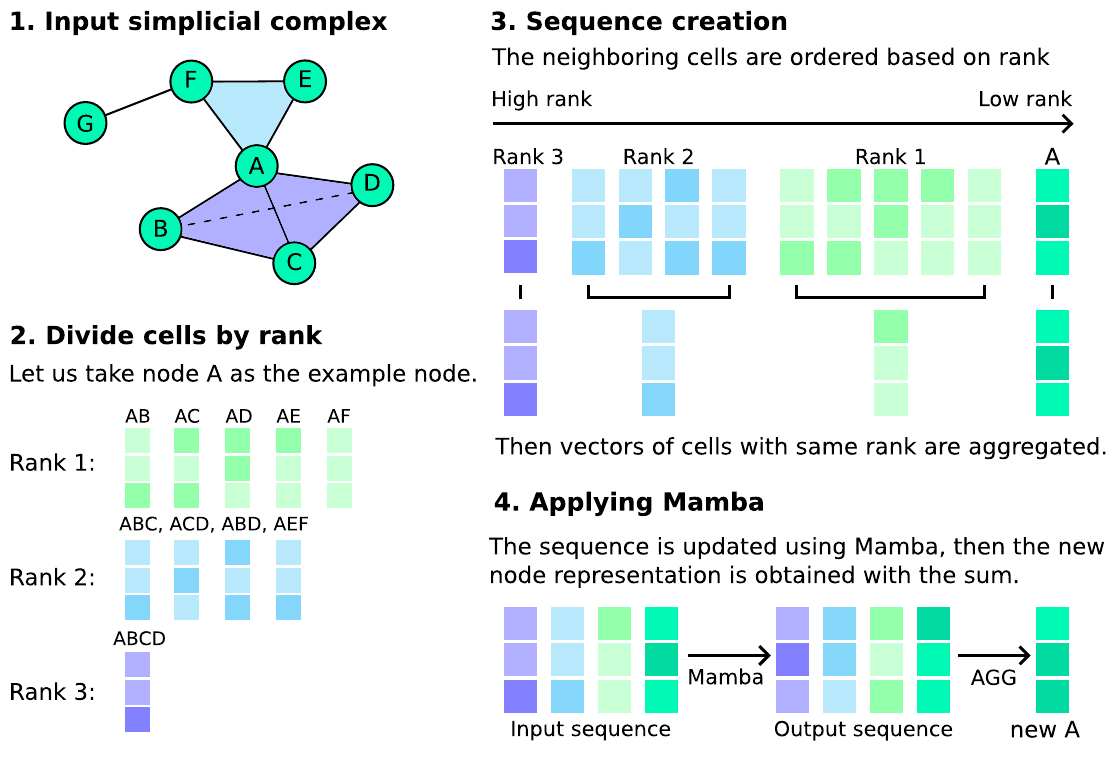}
    \caption{Scheme showing how our model updates the vector of node A. First, the neighboring cells of the target node are selected and then ordered by rank. A single vector for each rank is obtained by aggregating the cells with the same rank. Mamba is used on the resulting sequence which is then summed to obtain the new node representation. }
    \label{fig:sequencer}
\end{figure}
Recently State Space Models (SSMs) \cite{guEfficientlyModelingLong2022} have become a popular alternative to Transformer-based systems for sequence modeling. They can be seen as a combination of recurrent neural networks and convolutional neural networks \cite{guMambaLinearTimeSequence2023}, and their main advantage is that they have linear or non-linear scaling in the sequence length. By modeling long-range interactions with principled mechanisms \cite{guHiPPORecurrentMemory2020} they have achieved state-of-the-art results in long-range datasets, like the Long Range Arena \cite{tayLongRangeArena2020}. \\
Mamba \cite{guMambaLinearTimeSequence2023} has emerged as one of the most popular SSM. Its success can be attributed to two main factors: its ability to dynamically adapt parts of the parameters previously fixed in the SSMs equations based on input data, and its hardware-aware implementation, which enables linear scaling with sequence length. By making its parameters input-dependent, Mamba effectively filters out irrelevant information while retaining relevant details indefinitely.
Still, determining how to leverage graph structures to generate sequences suitable for SSMs is an area of active research \cite{behrouzGraphMambaLearning2024, dingRecurrentDistanceEncodingNeural2023a}. 
Topological Deep Learning (TDL) \cite{barbarossaTopologicalSignalProcessing2020a, hajijTopologicalDeepLearning2023c, papamarkouPositionPaperChallenges2024a} is a field of research also interested in addressing the limitations of graph neural networks. TDL studies topological spaces such as simplicial, cellular, and combinatorial complexes. These domains are not limited to relations incorporating only two nodes, and they can instead work with structures containing multiple. 
Adapting sequence models to work with topological domains presents several challenges. In graphs, the concept of neighborhood is straightforward, as nodes are neighbors if they share an edge. However, in topological structures, the notion of neighborhood is more complex. As described in Section \ref{Notation}, simplicial complexes are composed of cells, which have different ranks based on the number of nodes they contain. Each group of cells at a given rank has both an upper and a lower neighborhood. For a comprehensive discussion on the most common topological domains, we refer readers to \cite{papillonArchitecturesTopologicalDeep2024a}.
The expanded definition of neighborhoods in topological domains complicates the development of message-passing strategies. Consequently, finding optimal methods for propagating information through these structures remains an ongoing area of research.
\paragraph{Contribution of the paper.} We propose a novel approach that combines sequence modeling with topological deep learning. By creating a sequence of cells ordered by their rank, and then modeling it using the Mamba model, we can have information propagating from structures of one rank to any other in a single step.
To validate our approach, we perform experiments on different graph datasets by converting them to simplicial complexes. We compare our results to state-of-the-art models developed to work on simplicial complexes. We also discuss how our approach allows for efficient batching during training. The effects of this batching technique on memory usage, training time, and overall performance are evaluated through a series of experiments. This approach is also compared to a standard batching technique for simplicial models.

%% file: sections/2_related_works.tex
\section{Related Work}
Graph Neural Networks \cite{scarselliGraphNeuralNetwork2009} commonly rely on the message-passing mechanism \cite{gilmerNeuralMessagePassing2017}, where each node representation is updated by aggregating "messages" from its neighboring nodes. This process typically involves a weighted sum of the features of adjacent nodes. Consequently, two non-adjacent nodes can only influence each other through multiple message-passing steps. GNNs have been effectively applied to a wide range of tasks, including graph classification \cite{kipfSemiSupervisedClassificationGraph2017}, link prediction \cite{zhangLinkPredictionBased2018}, protein folding \cite{jumperHighlyAccurateProtein2021}, and recommendation systems \cite{wangKGATKnowledgeGraph2019}.\\
Recurrent Neural Networks (RNNs) are capable of capturing dependencies across sequences, which makes them suitable for problems involving temporal correlations \cite{salmanWeatherForecastingUsing2015, shiDeepLearningHousehold2018}. However, they face challenges such as vanishing gradients \cite{pascanuDifficultyTrainingRecurrent2013} and high computational costs due to their inherently sequential nature, which limits parallelization. State Space Models \cite{guEfficientlyModelingLong2022} provide a computationally efficient alternative to RNNs. The Mamba model \cite{guMambaLinearTimeSequence2023} has recently gained traction for its ability to offer a more flexible state representation while preserving computational efficiency.\\
This concept is leveraged in \cite{dingRecurrentDistanceEncodingNeural2023a} to develop a graph recurrent encoding-by-distance block, which creates sequences for each node by aggregating features from equidistant nodes, starting from the furthest nodes and ending with direct neighbors. This sequence can then be input into a recurrent neural network, providing an alternative to traditional message-passing architectures. Similarly, the authors in \cite{behrouzGraphMambaLearning2024} construct sequences by performing random walks of varying lengths from each node and ordering these walks by their lengths, allowing for the generation of arbitrarily long sequences.\\
Building on these ideas, we explore the creation of sequences when working with simplicial complexes. Topological Deep Learning \cite{barbarossaTopologicalSignalProcessing2020a, hajijTopologicalDeepLearning2023c, papamarkouPositionPaperChallenges2024a} is increasingly recognized for its capacity to model complex relationships among graph nodes, leveraging various topological domains, including simplicial complexes, cellular complexes, hypergraphs, and combinatorial complexes \cite{papillonArchitecturesTopologicalDeep2024a}.\\
Many models have been developed to work directly with simplicial complexes. In the SCN model presented in \cite{wuSimplicialComplexNeural2024} the 0-cell representations are calculated by aggregating the 1-cell and the 2-cell representations by using the adjacency between nodes and edges and nodes and triangles, and by adding a learnable weight matrix. The same is done for the edge and the triangle representations, effectively having the cells of each rank influence every other rank directly. The main limitation of this approach is that it considers only 0, 1, and 2-cells, while simplicial complexes can in theory consider cells of any arbitrary rank. 
In \cite{yangConvolutionalLearningSimplicial2023} the authors propose the SCCNN model that learns the representations of the simplices by performing convolutions considering lower and upper adjacencies independently. While this approach can be applied to cells of any rank, when cells differ in rank by more that one, they no longer directly influence each other. 
As it will be presented shortly our work directly tackles both limitations presented by introducing a model in which cells of any rank directly influence the cells of any other rank. 

%% file: sections/3_method.tex
\section{Notation and Background} \label{Notation}
A graph $\mathcal{G}=(V, E)$ is a tuple composed of a finite set of nodes $V$ and a finite set of edges $E$ connecting pairs of nodes. Considering a graph with $n_0$ nodes, $n_1$ edges, node features of dimension $d_0$, and edge features with dimension $d_1$, we obtain a featured graph $\mathcal{G}_{F}=(V, E, F_V, F_E)$ where $F_V: V \to \mathbb{R}^{d_0}$ and $F_E: E \to \mathbb{R}^{d_1}$ are functions that map the nodes and the edges to their feature vectors respectively. One common choice to encode the structure of the graph is to use the adjacency matrix $A\in\mathbb{R}^{n_0 \times n_0}$ where 
\begin{equation*}
    (A)_{i,j} = 
    \begin{cases}
        1 & e_{ij} \in E, \\
        0 & \text{otherwise}, \\
    \end{cases}
\end{equation*}
with $e_{ij}$ representing an edge between nodes $i$ and $j$. \\
Simplicial complexes expand on graphs by considering cells with different numbers of nodes. 
\begin{definition} \label{def:simplicial}
A \emph{simplicial complex} is a tuple $\mathcal{X}=(\mathcal{X}_0, \mathcal{X}_1, ..., \mathcal{X}_K)$ of finite ordered sets, where $\mathcal{X}_0$ is a set of nodes (or 0-cells), $\mathcal{X}_1$ is a set of edges (or 1-cells) with each edge $e\in \mathcal{X}_1$ being an ordered pair of nodes $e=[v_1, v_2]$ for $v_1, v_2 \in \mathcal{X}_0$. Similarly, $\mathcal{X}_2$ is a set of triangles (2-cells) with each face $\sigma \in \mathcal{X}_2$ being an ordered sequence of edges $\sigma = [e_1, e_2, e_3]$ forming a closed path. $\mathcal{X}_k$ is a set of $k$-cells, where each is an ordered sequence of cells of rank $k-1$. A \emph{featured simplicial complex} is a tuple $\mathcal{X}_F=(\mathcal{X}_0, \mathcal{X}_1, ..., \mathcal{X}_K, F_{\mathcal{X}_0}, F_{\mathcal{X}_1}, ..., F_{\mathcal{X}_K})$ where $F_{\mathcal{X}_r}: \mathcal{X}_r \to \mathbb{R}^{d_r}, ~ \forall r \in [0, 1, ..., K]$ maps each cell in $\mathcal{X}_r$ to a feature vector in $\mathbb{R}^{d_r}$.
\end{definition}
We use $n_i$ to indicate the number of cells of rank $i$ in the simplicial complex, and  $h_x^{(r)}$ to indicate the features of simplex $x$, which has rank $r$. $H^{(r)} \in \mathbb{R}^{n_r \times d_r}$ is the matrix where each row contains the features of a cell of rank $r$ in the simplex.

Cell $x^{(r_1)}$ of rank $r_1$ is on the \textit{boundary} of cell $y^{(r_2)}$ of rank $r_2$ if $x$ is connected to $y$ and $r_1<r_2$, and we express this relation as $x^{(r_1)} \prec y^{(r_2)}$, as presented in \cite{papillonArchitecturesTopologicalDeep2024a}.
We denote as $B_r \in \mathbb{R}^{n_{r-1} \times n_r}$ the unweighted incidence matrices defined as
\begin{equation*}
    (B_r)_{i,j} = 
    \begin{cases}
        \pm 1 & x_i^{(r-1)} \prec x_j^{(r)}, \\
        0 & \text{otherwise}. \\
    \end{cases}
\end{equation*}
The sign of the entries of $B_r$ encodes the orientation of the cells of the simplicial complex. 
\subsection{Clique lifting} 
While higher-order data naturally emerge in a variety of systems, from social networks \cite{knokeSocialNetworkAnalysis2008} to proteins \cite{ jhaPredictionProteinProtein2022}, the complexity of accurately capturing these interactions limits the collection of such data. Consequently, higher-order data are frequently derived by transforming graph datasets and augmenting them with higher-order features.
Following the literature \cite{telyatnikovTopoBenchmarkXFrameworkBenchmarking2024} we call \textit{lifting} the transformation of a graph into a simplicial complex.  
\begin{definition}
\label{def:lifting}
Let \(\mathcal{G}_{F} = (V, E, F_V, F_E)\) be a featured graph. A lifting of $\mathcal{G}_{F}$ is a triplet $(\mathcal{X}_{F},\iota, L)$, where
$\mathcal{X}_{F} = (\mathcal{X}_0, \mathcal{X}_1, ..., \mathcal{X}_K, F_{\mathcal{X}_0}, F_{\mathcal{X}_1}, ..., F_{\mathcal{X}_K}),$
is a featured simplicial complex, $\iota$ is an embedding map that maps the nodes and edges of $\mathcal{G}$ to cells in $\mathcal{X}_{F}$, and $L$ is a collection of functions that map features of $\mathcal{G}_{F}$ to features of $\mathcal{X}_{F}$. The embedding map $\iota$ satisfies $\iota(v) = s_v$ for $v \in V$ with $s_v \in \mathcal{X}_0$ being a $0$-cell corresponding to node $v$, and $\iota(e) = x_e$ for $e \in E$ with $x_e \in \mathcal{X}_1$ being a cell corresponding to edge $e$. The collection of functions in $L$ consists of a lifting procedure for the higher-order cells in $\mathcal{X}_{F}$ that are not part of $\mathcal{G}$ and a lifting procedure for the feature maps $\{F_{\mathcal{X}^i}\}_{i=0}^{K}$ 
based on $\mathcal{G}_{F}$.
\end{definition}

\paragraph{Connectivity lifting} The lifting chosen is the \textit{clique lifting}: given a graph $\mathcal{G}$ and a maximum rank $R$ this lifting first finds the set of cliques $\mathcal{C}$ of the graph. Then for each clique $C \in \mathcal{C}$ the following sets of cells are added to the simplicial complex
\begin{equation*}
    \mathcal{X}_r = \bigcup_{C \in \mathcal{C}} \left\{ X \subseteq C ~|~ |X|=r \right\}, ~ \forall 1 \leq r \leq R.
\end{equation*}
This approach has the advantage of respecting the initial connectivity of the graph since the $1$-rank cells of the obtained simplicial complex will be all the edges of the original graph but no pair of unconnected nodes will be included in the $1$-cells.
It is important to notice that given a clique of $n$ elements, the number of rank $r$ cells that the algorithm creates is the binomial coefficient $\binom{n}{r+1}$ which can lead to an exploding number of high-rank cells when the dataset presents large cliques. Despite this, the lifting works well in all the datasets considered in our experiments and we leave an exploration of other lifting techniques to future work.

\paragraph{Feature lifting} Since the obtained higher-order cells do not have predefined features, it is necessary to \textit{lift} the features as well. Many approaches are possible, using both fixed or learnable functions, but the most common approach is to obtain the features of a cell by simply summing the features of the lower rank cells that compose it. 
When using the sum, obtaining the features of the cells of any rank is straightforward when the features of the lower-rank cells are known, since we can use
\begin{equation}
\label{eq:feature_lift}
H^{(r)} = B_r^\top H^{(r-1)}.
\end{equation}

\section{Method}
This section presents in detail the proposed model architecture, which we refer to as \textit{TopoMamba}, and then discusses batching for simplicial complexes, starting with presenting the general approach and then detailing how our model architecture allows for a more efficient batching implementation.

\subsection{Model architecture} \label{model_architecture}
TopoMamba consists of three parts that will be explained in the following paragraphs: (i) the feature encoder, (ii) the Mamba block, and (iii) the task head. 
\paragraph{Feature encoder} The feature encoder performs the transformation of the input features via an initial mapping function
\begin{equation*}
    H^{(0)} = \text{ReLU} \left( H_{in}^{(0)} \cdot W_{in}^\top + b_{in} \right),
\end{equation*}
where $H_{in}^{(0)} \in \mathbb{R}^{n \times d_{in}}$ is the matrix of input features for the $n$ nodes of the simplicial complex. The output of the feature encoder is $H^{(0)} \in \mathbb{R}^{n \times d_h}$, where $d_h$ is the hidden dimension of the transformed features. $W_{in} \in \mathbb{R}^{d_h \times d_{0}}$ is a learnable weight matrix, and $b_{in} \in R^{d_h}$ is a learnable bias vector. \\

\paragraph{Mamba block} The Mamba model takes as input a batch of sequences and outputs their updated representation. To be able to use Mamba we construct a sequence for each node from the simplicial complex structure. We use a parameter-free transformation (see Figure \ref{fig:sequencer} steps 2, 3). In particular, for a given node $A$, we collect all the simplices it belongs to and divide them by rank: we obtain $R$ sets of the form 
\begin{equation*}
Z_A^{(r)} = \{h^{(r)} ~ | ~ A \prec x^{(r)} \},
\end{equation*}
with $R$ the maximum rank of the simplicial complex. The features for the higher-order cells are obtained from the node features using Equation \ref{eq:feature_lift}. Next, the collected simplices of the same rank are aggregated using the AGG function, which pools a set of $i$-cells into a single representation. The AGG function can be any invariant aggregation operation, as there is no natural order among the cells of the same rank.
The obtained sequence for node $A$ is
\begin{equation} \label{eq:sequence}
    \mathcal{S}_{A} = \left[ \text{AGG}(Z_A^{(R)}),~\text{AGG}(Z_A^{(R)}), ...,~\text{AGG}(Z_A^{(1)}),~h_{A}^{(0)} \right].
\end{equation}
The aggregation operation employed in our model is the sum, since it has better expressive power than the mean and the max aggregators \cite{xuHowPowerfulAre2019}.
The features of the node being considered are also added as the last element of the sequence. This ensures that the model can distinguish between different nodes with similar neighborhoods. The obtained sequences are then passed to the Mamba model $\mathcal{M}$ which outputs new sequences
\begin{equation}
\mathcal{S}_{A}^1 = \mathcal{M}(\mathcal{S}_{A}).  
\end{equation}
We also employ a separate Mamba model and apply it to the inverse of the sequence. This makes it possible to have the features of the higher-order cells be influenced by the node being studied, since as pointed out before the node vector is the last element. If we indicate with $\text{inv}(\cdot)$ the operator that inverts a sequence, then we can define
\begin{equation}
\mathcal{S}_{A}^2 = \mathcal{M}(\text{inv}(\mathcal{S}_{A})). 
\end{equation}
The results of the two Mamba layers are then summed to obtain a single sequence
\begin{equation} \label{eq:inv_mamba}
    \mathcal{S}_{A} = \mathcal{S}_{A}^1 + \text{inv}(\mathcal{S}_{A}^2).
\end{equation}
This equation demonstrates how cells of different ranks can directly influence one another. In sequential models, earlier elements in a sequence affect those that follow. Thus, in $\mathcal{S}_A^1$, higher-order cells influence lower-rank cells, while in $\mathcal{S}_A^2$, the influence is reversed, with lower-rank cells affecting higher-rank ones. This enables interactions across all ranks within a single Mamba block.
The new features for node $A$ are then obtained by summing the elements of the newly obtained sequence
\begin{equation}
    \tilde h_{A}^{(0)} = \sum \mathcal{S}_{A}.
\end{equation}
In parallel to this whole block, we also employ skip connections which were shown to improve the model performance (Section \ref{results}). The final node features can be written as 
\begin{equation} \label{eq:mamba_block_output}
h_{A}^{(0)} \leftarrow h_{A}^{(0)} + \tilde h_{A}^{(0)}.
\end{equation}
Equations \ref{eq:sequence} to \ref{eq:mamba_block_output} form the Mamba block. This module outputs the node features but does not update the higher-order cells, which can be obtained by applying Equation \ref{eq:feature_lift} again. This allows us to stack Mamba blocks one after the other, obtaining a final $h_A^{(0)}$. By exploiting the parallelization capabilities of the Mamba model we can stack the sequences for each node into a single matrix and obtain the desired output matrix $H^{(0)}$.
\\ 
\paragraph{Task head} Finally, the task head is another linear layer that transforms the hidden representations for each node into the shape needed for the desired task. In particular
\begin{equation*}
    H_{out}^{(0)} = \text{ReLU} \left( H^{(0)} \cdot W_{out}^\top + b_{out} \right),
\end{equation*}
where $H_{out} \in \mathbb{R}^{n \times d_{out}}$ is the final output of the network, $W_{out} \in \mathbb{R}^{d_{out} \times d_h}$ is the learnable weight matrix for the transformation, and $b_{out} \in \mathbb{R}^{d_{out}}$ is the bias vector. For classification $d_{out}$ corresponds to the number of possible classes, while $d_{out}=1$ is used for regression. \\
For clarity, we omitted from the previous discussions the inclusion of dropout layers and normalization layers. In particular, dropout is implemented after both the feature encoder and the task head, while layer normalization is used after creating the sequences.

\subsection{Batching} \label{node_incidence}
Batching is essential for reducing memory requirements when working with large datasets. Unlike standard machine learning, batching for relational data requires careful handling to avoid loss of information. In message-passing networks, the limited spread of information, restricted by the number of message-passing steps, is leveraged by sampling algorithms. In fact, to avoid losing network information, they select only the batch nodes along with their corresponding n-hop neighborhoods. In TDL, the challenge of handling large higher-order networks becomes more pronounced due to the inclusion of higher-order structures. Generally, since the neighborhood structures of simplicial complexes are more complicated than those in graphs, it is required to handle each incidence matrix $B_r$ independently. This involves first removing nodes that are too distant, followed by edges, triangles, and so on for each higher-order structure. While this approach can be effective, it requires a series of matrix operations, which slows down model training.

TopoMamba allows us to address the batching issue just mentioned by a sampling mechanism that is conceptually similar to, and just as straightforward as, graph neighborhood batching.
For the creation of the sequences needed as input to the Mamba block, it is enough to rely solely on the boundary relations between higher-order structures and nodes, referred to as $B^*$, defined as:
\begin{definition}
Let $B^* \in \mathcal{R}^{n_0 \times n^*}$ be a matrix with $n^*= \sum_{i=1}^R n_i$ the sum of the number of all the higher order structures starting from rank 1 to the maximum rank. The \emph{node incidence} is the matrix
\begin{equation*}
    (B^*)_{i,j} = 
    \begin{cases}
        \pm 1 & x_i^{(0)} \prec x_j^{(r)}, \\
        0 & \text{otherwise}. \\
    \end{cases}
\end{equation*}
The rank $r$ can be any $1 \leq r \leq R$.
\end{definition}

By utilizing the node incidence matrix, we can directly apply existing neighborhood sampling algorithms developed for graphs to our model. In simplicial complexes, nodes that are part of a higher-order structure are all connected among themselves (see Definition \ref{def:simplicial}). This ensures that when performing neighborhood sampling, all the cells belonging to the higher-order structures adjacent to a node are included in the selection. This method is efficient because it relies on a single incidence matrix and enables the use of well-optimized, graph-based libraries.

%% file: sections/4_experiments.tex
\section{Empirical Analysis}
In this section, we present a comprehensive evaluation of our proposed model, comparing its performance against state-of-the-art models for simplicial complexes. To ensure generalizability, we evaluate the models on datasets from diverse domains. Section \ref{setup} outlines the experimental setup, while Section \ref{results} discusses the main results, examines the impact of batching on the proposed and simplicial models, and includes an ablation study to analyze the effects of the model components.
\subsection{Setup}
\label{setup}
We use several datasets in this study, including Cora~\cite{mccallumAutomatingConstructionInternet2000}, Citeseer~\cite{gilesCiteSeerAutomaticCitation1998}, Pubmed~\cite{senCollectiveClassificationNetwork2008}, Minesweeper, Amazon Ratings, Roman Empire~\cite{platonovCriticalLookEvaluation2022}, and US-county-demos~\cite{jiaResidualCorrelationGraph2020}. Table \ref{tab:data_statistics} provides key statistics for each dataset, where the numbers of triangles and tetrahedra correspond to the datasets after their transformation using clique lifting.
Cora, Citeseer, and Pubmed are well-known cocitation datasets commonly used to benchmark graph neural networks for node classification tasks. Minesweeper, Amazon Ratings, and Roman Empire are recent heterophilious datasets, characterized by their non-homophilous structure, which poses a challenge for message-passing networks \cite{platonovCriticalLookEvaluation2022}. US-county-demos is a regression dataset where any node attribute can serve as the target label \cite{jiaResidualCorrelationGraph2020}.
All datasets are randomly split into training, validation, and test sets, with $50\%$ for training and $25\%$ each for validation and testing. The results are averaged over ten randomly initialized runs. For classification tasks, accuracy is used as the evaluation metric, except for the Minesweeper dataset, where we follow the approach in~\cite{platonovCriticalLookEvaluation2022} and use the ROC-AUC metric. For regression tasks, including the US-Birth and US-Migration datasets, we use the mean absolute error as the evaluation metric.

All models are optimized using the Adam optimizer~\cite{kingmaAdamMethodStochastic2014}. A hyperparameter search is performed to identify the optimal settings for each dataset. Training is monitored using early stopping, with the best parameters selected based on validation performance. The hyperparameter search is executed on a Linux machine with 32 cores, 256GB of system RAM,
and one NVIDIA A6000 GPU, with 48GB of GPU memory.
\input{sections/tables/data_statistics}

%% file: sections/tables/data_statistics.tex

\begin{table}[h]
\centering
\begin{adjustbox}{width=1.\textwidth}
\begin{tabular}{lccccc}
\toprule
                & \# of nodes & \# of node features & \# of edges & \# of triangles & \# of tetrahedra \\ 
\midrule
Cora            & 2708        & 1433                & 5278        & 1630            & 220              \\
Citeseer        & 3327        & 3703                & 4552        & 1167            & 255              \\
Pubmed          & 19717       & 500                 & 44324       & 12520           & 3275             \\
Minesweeper     & 10000       & 7                   & 39402       & 39402           & 9801             \\
Roman           & 22662       & 300                 & 7168        & 0               & 0                \\
Amazon          & 24492       & 300                 & 93050       & 110765          & 64195            \\
US-county-demos & 3224        & 6                   & 9483        & 6490            & 225              \\  
\bottomrule
\end{tabular}
\end{adjustbox}
\caption{Statistics for the different datasets. The number of higher-order simplices is reported for the clique expansion algorithm. It is worth noticing that because of the peculiar structure of the Roman dataset, there are no cliques of three or more nodes, so the simplex obtained is just a graph.}
\label{tab:data_statistics}
\end{table}

%% file: sections/5_results.tex
\subsection{Experiments}
\label{results}

\paragraph{Models comparison}
To assess our model's performance, we compare it with two state-of-the-art models for simplicial complexes: SCN~\cite{wuSimplicialComplexNeural2024} and SCCNN~\cite{yangConvolutionalLearningSimplicial2023}. We also test a variant of our model that substitutes the Mamba backbone with a recurrent neural network based on gated recurrent units~\cite{choLearningPhraseRepresentations2014}.

Table \ref{tab:comparison} presents the performance results for the models evaluated in our experiments. TopoMamba generally performs better or is comparable to state-of-the-art simplicial models across all datasets, except for the Roman Empire dataset. The fact that our model relies solely on graph structures to generate sequences for the Mamba model highlights its ability to effectively capture and leverage topological information. However, the Roman Empire dataset poses a challenge due to its chain-like structure, characterized by an average degree of 2.9 and a diameter of 6824, which makes it difficult for our model to learn effective node representations.

\paragraph{Batching}
The three models TopoMamba, SCN, and SCCNN are compared both when batching and when using the full-batch approach, where the entire graph is used for training. For our model the batching technique introduced in Section \ref{node_incidence} is used. Since it cannot be adapted for the other models, with them we employ the standard batching approach also described in the same section.  
In addition to performance, we measure the maximum memory usage during training and the time required for each epoch. 

Training times per epoch are reported in Table \ref{tab:time}. As expected, the batching approach tends to result in longer training times compared to full-batch training. However, it is important to note that the model typically requires fewer training epochs when using batching. Our model also consistently demonstrates faster training times compared to other models while achieving comparable or superior performance.

Additional metrics are detailed in Table \ref{tab:batch_size}. The results suggest that there is no optimal batching method; for certain datasets, batching offers performance advantages, while for others, full-batch training is more effective. In terms of memory usage, the results confirm that our batching method is an effective technique for enabling large-scale dataset processing with limited memory resources.

\paragraph{Ablation Study: Effect of Model Components}
An ablation study is also performed to understand the impact of the various components of TopoMamba, including the presence or absence of skip connections and the backward Mamba model, as well as variations in hidden size and the number of blocks. 
The findings from the ablation study are summarized in Table \ref{tab:ablation}. The results indicate that using two layers and a larger hidden dimension generally improves performance. As anticipated, the inclusion of skip connections is particularly beneficial when increasing the number of layers. Furthermore, considering the standard deviation in the results, our model appears robust to minor changes in architecture and hyperparameters.

\input{sections/tables/comparison}

\input{sections/tables/ablation}

\input{sections/tables/time}

%% file: sections/tables/comparison.tex
\begin{table}[ht]
\centering
\begin{tabular}{lcccc}
\toprule
Datasets & TopoMamba & RNN & SCCNN & SCN   \\ 
\midrule
Cora             & $\mathbf{86.50 \pm 1.03}$ & $84.87 \pm 0.89$ & \cellcolor{blue!25}$85.58 \pm 1.53$ & $84.71 \pm 0.85$ \\
Citeseer         & $\mathbf{73.96 \pm 1.62}$ & $72.53 \pm 1.47$ & $73.12 \pm 1.62$ & \cellcolor{blue!25}$73.36 \pm 1.91$ \\
Pubmed           & \cellcolor{blue!25}$88.44 \pm 0.48$ & $88.20 \pm 0.50$ & $88.18 \pm 0.32$ & $\mathbf{88.72 \pm 0.50}$ \\
Minesweeper      & \cellcolor{blue!25}$89.19\pm0.61$ & $88.87 \pm 0.59$ & $89.02 \pm 0.20$ & $\mathbf{90.32 \pm 0.11}$ \\
Roman      & $82.74 \pm 0.43$ & $80.79 \pm 0.60$ & $\mathbf{89.15 \pm 0.32}$ & \cellcolor{blue!25}$88.79 \pm 0.46$ \\
Amazon      & $\mathbf{48.22 \pm 1.01}$ & \cellcolor{blue!25}$47.21 \pm 0.86$ & $45.10\pm0.74$ & $43.52\pm1.40$ \\
\midrule
US Birth         & $\mathbf{0.63 \pm 0.02}$ & $0.77 \pm 0.04$ & \cellcolor{blue!25}$0.64 \pm 0.04$ & $0.71 \pm 0.08$ \\
US Migration     & $\mathbf{0.58 \pm 0.03}$ & $0.77 \pm 0.05$ & \cellcolor{blue!25}$0.60 \pm 0.06$ & $0.75 \pm 0.04$ \\
\bottomrule
\end{tabular}
\caption{Comparison between the different models. The results are shown as mean and standard deviation over 10 runs. The best results are in bold, while the second best are highlighted in blue. For \textit{US Birth} and \textit{US Migration} lower is better since the metric reported is the mean absolute error.}
\label{tab:comparison}
\end{table}

%% file: sections/tables/ablation.tex
\begin{table}
\centering
\begin{tabular}{cccccc}
\toprule
\multicolumn{1}{l}{} & \multicolumn{1}{l}{} & \multicolumn{2}{c}{Skip connection}  & \multicolumn{2}{c}{No skip connection} \\
\# layers            & $d_h$     & With $\mathcal{S}_A^2$ & Without $\mathcal{S}_A^2$ & With $\mathcal{S}_A^2$ & Without $\mathcal{S}_A^2$  \\ \midrule
\multirow{2}{*}{1}   & 128                  & $84.55\pm1.53$ & $84.33\pm1.74$    & $84.71\pm1.30$  & $84.36\pm1.30$     \\
                     & 256                  & $85.36\pm1.32$ & $85.81\pm1.23$    & $85.35\pm1.37$  & $85.54\pm1.00$     \\ \midrule
\multirow{2}{*}{2}   & 128                  & $85.32\pm1.36$ & $86.20\pm1.10$    & $85.95\pm1.09$  & $85.16\pm1.16$     \\ 
                     & 256                  & $86.50\pm1.03$ & $86.15\pm1.64$    & $85.57\pm1.11$  & $85.45\pm1.14$     \\
\bottomrule
\end{tabular}
\caption{Ablation study on the Cora dataset. \textit{\# layers} refers to the number of Mamba blocks used, while $S_A^2$ refers to using or not Mamba on the inverse of the sequence as presented in Equation \ref{eq:inv_mamba}.}
\label{tab:ablation}
\end{table}

%% file: sections/tables/time.tex
\begin{table}[ht]
\centering
\begin{tabular}{ccccc}
\toprule
Dataset                       & Batch Size & TopoMamba [s] & SCN [s]& SCCNN [s] \\
\midrule
\multirow{2}{*}{Cora}         &  128       & 11 & 29 & 25  \\
                              & full       & 10 & 28 & 32   \\ \midrule
\multirow{2}{*}{Citeseer}     & 128        & 12 & 18 & 21    \\
                              & full       & 10 & 21 & 32    \\ \midrule
\multirow{2}{*}{Pubmed}       & 128        & 116 & 414 & 508   \\
                              & full       & 22 & 286 & 367    \\ \midrule
\multirow{2}{*}{Minesweeper}  & 128        & 48 & 226 & 184    \\
                              & full       & 31 & 409 & 771    \\ \midrule
\multirow{2}{*}{Roman}        & 128        & 48 & 90 & 242    \\
                              & full       & 109 & 100 & 332   \\ \midrule
\multirow{2}{*}{Amazon}       & 128        & 633 & 2477 & 3186   \\
                              & full       & 53 & 217 & 241    \\ \midrule
\multirow{2}{*}{US Birth}     & 128        & 14 & 79 & 84    \\
                              & full       & 19 & 212  & 115    \\ \midrule
\multirow{2}{*}{US Migration} & 128        & 16 & 48 & 47    \\
                              & full       & 15 & 114 & 94    \\
\bottomrule
\end{tabular}
\caption{Table showing the training times in seconds for an epoch. The results are reported for the studied models both when batching is used and when the whole graph is passed to the model. }
\label{tab:time}
\end{table}

%% file: sections/6_conclusions.tex
\section{Conclusions}
This paper introduces a novel architecture for processing simplicial complexes. Our approach first constructs a sequence for each node by ordering and aggregating the representations of neighboring cells according to their rank. Then the Mamba model is used to iteratively update the representations of the nodes in the simplicial complex. The proposed approach eliminates the need to devise a specific higher-order message-passing schema, as it allows cells of different ranks to communicate directly (see Section \ref{model_architecture}). The effectiveness of our model is evaluated against several state-of-the-art networks across various datasets, demonstrating that it either outperforms or matches these models while TopoMamba is considerably faster (see Section \ref{results}). Furthermore, the proposed architecture allows for a novel batching strategy relying on the introduced concept of node incidence (see Section \ref{node_incidence}) that is easy to implement, with minimal computational overhead during training, thereby enhancing the scalability of the approach. 

For future work, TopoMamba can be extended to other topological domains, offering a novel paradigm for information propagation across different structures and potentially addressing the challenge of the increased time complexity in topological neural networks compared to GNNs. For example, this approach could be adapted to cellular and combinatorial complexes, which also exhibit hierarchical relationships among their constituent cells. Additionally, the node incidence matrix introduced in this work is analogous to the incidence matrix used for hypergraphs, suggesting that TopoMamba could be adapted to operate effectively in this domain as well.
Additionally, Mamba can potentially allow the development of a single architecture for multiple topological domains (see open problem 6 \cite{papamarkouPositionPaperChallenges2024a}), by combining in a single sequence multiple topological domains. 

%% file: sections/appendix.tex
\section{Appendix}
\input{sections/tables/batch_size}

%% file: sections/tables/batch_size.tex
\begin{table}[ht]
\centering
\begin{adjustbox}{width=0.84\textwidth}
\begin{tabular}{llcccc}
\toprule
Dataset                       & Model      & Batch Size & Results         & GPU Memory [Mb] & Epoch Time [s]\\
\midrule
\multirow{6}{*}{Cora}         & \multirow{2}{*}{TopoMamba} & 128        & $84.93\pm1.22$  & 400        & 11    \\
                              &                            & full       & $85.08\pm1.53$  & 639        & 10    \\ \cmidrule{2-6} 
                              & \multirow{2}{*}{SCN}       & 128        & $84.62\pm1.46$  & 242        & 29    \\
                              &                            & full       & $84.71\pm0.85$  & 404        & 28    \\ \cmidrule{2-6} 
                              & \multirow{2}{*}{SCCNN}     & 128        & $85.06\pm1.51$  & 238        & 25    \\
                              &                            & full       & $85.58\pm1.53$  & 404        & 32    \\
\midrule
\multirow{6}{*}{Citeseer}     & \multirow{2}{*}{TopoMamba} & 128        & $72.50\pm1.33$  & 309        & 12    \\
                              &                            & full       & $73.96\pm1.62$  & 757        & 10    \\ \cmidrule{2-6} 
                              & \multirow{2}{*}{SCN}       & 128        & $72.71\pm1.29$  & 389        & 18    \\
                              &                            & full       & $73.36\pm1.91$  & 389        & 21    \\ \cmidrule{2-6} 
                              & \multirow{2}{*}{SCCNN}     & 128        & $72.81\pm0.95$  & 389        & 21    \\
                              &                            & full       & $73.12\pm1.62$  & 389        & 32    \\
\midrule
\multirow{6}{*}{Pubmed}       & \multirow{2}{*}{TopoMamba} & 128        & $88.28\pm0.59$  & 1546       & 116   \\
                              &                            & full       & $87.61\pm0.39$  & 4507       & 22    \\ \cmidrule{2-6} 
                              & \multirow{2}{*}{SCN}       & 128        & $87.20\pm0.46$  & 1520       & 414   \\
                              &                            & full       & $86.59\pm0.47$  & 2185       & 286   \\ \cmidrule{2-6} 
                              & \multirow{2}{*}{SCCNN}     & 128        & $87.37\pm0.59$  & 1464       & 508   \\
                              &                            & full       & $86.16\pm0.63$  & 2194       & 367   \\
\midrule
\multirow{6}{*}{Minesweeper}  & \multirow{2}{*}{TopoMamba} & 128        & $89.19\pm0.61$  & 665        & 48    \\
                              &                            & full       & $85.51\pm0.89$  & 2296       & 31    \\ \cmidrule{2-6} 
                              & \multirow{2}{*}{SCN}       & 128        & $85.77\pm0.83$  & 691        & 226   \\
                              &                            & full       & $83.02\pm0.86$  & 2402       & 409   \\ \cmidrule{2-6} 
                              & \multirow{2}{*}{SCCNN}     & 128        & $86.21\pm0.80$  & 688        & 184   \\
                              &                            & full       & $88.01\pm0.92$  & 2414       & 771   \\
\midrule
\multirow{6}{*}{Roman}        & \multirow{2}{*}{TopoMamba} & 128        & $82.55\pm0.74$  & 488        & 48    \\
                              &                            & full       & $80.59\pm0.93$  & 1098       & 109   \\ \cmidrule{2-6} 
                              & \multirow{2}{*}{SCN}       & 128        & $67.65\pm1.31$  & 113        & 90    \\
                              &                            & full       & $70.25\pm1.22$  & 1151       & 100   \\ \cmidrule{2-6} 
                              & \multirow{2}{*}{SCCNN}     & 128        & $76.47\pm0.84$  & 113        & 242   \\
                              &                            & full       & $76.83\pm1.12$  & 1161       & 332   \\
\midrule
\multirow{6}{*}{Amazon}       & \multirow{2}{*}{TopoMamba} & 128        & $47.88\pm1.16$  & 2322       & 633   \\
                              &                            & full       & $49.60\pm0.47$  & 24568      & 53    \\ \cmidrule{2-6} 
                              & \multirow{2}{*}{SCN}       & 128        & $43.52\pm1.40$  & 2669       & 2477  \\
                              &                            & full       & $43.42\pm1.32$  & 24887      & 217   \\ \cmidrule{2-6} 
                              & \multirow{2}{*}{SCCNN}     & 128        & $45.10\pm0.74$  & 2597       & 3186  \\
                              &                            & full       & $44.96\pm.1.12$ & 24807      & 241   \\
\midrule
\multirow{6}{*}{US Birth}     & \multirow{2}{*}{TopoMamba} & 128        & $0.66\pm0.02$  & 350         & 14    \\
                              &                            & full       & $0.88\pm0.04$  & 580         & 19    \\ \cmidrule{2-6} 
                              & \multirow{2}{*}{SCN}       & 128        & $0.75\pm0.02$  & 855         & 79    \\
                              &                            & full       & $0.90\pm0.07$  & 855         & 212   \\ \cmidrule{2-6} 
                              & \multirow{2}{*}{SCCNN}     & 128        & $0.64\pm0.04$  & 815         & 84    \\
                              &                            & full       & $1.15\pm0.59$  & 815         & 115   \\
\midrule
\multirow{6}{*}{US Migration} & \multirow{2}{*}{TopoMamba} & 128        & $0.59\pm0.03$  & 350         & 16    \\
                              &                            & full       & $1.11\pm0.05$  & 580         & 15    \\ \cmidrule{2-6} 
                              & \multirow{2}{*}{SCN}       & 128        & $0.75\pm0.04$  & 855         & 48    \\
                              &                            & full       & $0.83\pm0.02$  & 855         & 114   \\ \cmidrule{2-6} 
                              & \multirow{2}{*}{SCCNN}     & 128        & $0.60\pm0.06$  & 815         & 47    \\
                              &                            & full       & $1.34\pm0.03$  & 815         & 94    \\
\bottomrule
\end{tabular}
\end{adjustbox}
\caption{Table showing the results from the experiments comparing using full batches or batch size of 128. For the US Birth and US Migration datasets lower is better since the metric reported is the mean absolute error.}
\label{tab:batch_size}
\end{table}